\pdfoutput=1

\documentclass[11pt]{article}

\usepackage{acl}

\usepackage{times}
\usepackage{latexsym}
\usepackage{multirow}
\usepackage{booktabs}
\usepackage{float}
\usepackage[T1]{fontenc}

\usepackage[utf8]{inputenc}

\usepackage{microtype}

\usepackage{twemojis}
\usepackage{xcolor}
\usepackage{soul}

\usepackage{fdsymbol}

\title{CLIMB \texttwemoji{climbing} -- Curriculum Learning for Infant-inspired Model Building}

\newcommand*\samethanks[1][\value{footnote}]{\footnotemark[#1]}

\author{
    {\bf Richard Diehl Martinez} \texttwemoji{flying_saucer} ~~~~
    {\bf Z\'{e}bulon Goriely} \texttwemoji{flying_saucer}\thanks{~ Equal contribution} ~~~~
    {\bf Hope McGovern} \texttwemoji{flying_saucer}\samethanks \\
    {\bf Christopher Davis} \texttwemoji{flying_saucer}\texttwemoji{sailboat} ~~~~
    {\bf Andrew Caines} \texttwemoji{flying_saucer}\texttwemoji{sailboat} ~~~~
    {\bf Paula Buttery} \texttwemoji{flying_saucer}\texttwemoji{sailboat} ~~~~
    {\bf Lisa Beinborn} \texttwemoji{herb} \\
    \texttwemoji{flying_saucer} Department of Computer Science \& Technology, University of Cambridge, U.K. \\
    \texttwemoji{sailboat} ALTA Institute, University of Cambridge, U.K. \\
    \texttwemoji{herb} Vrije Universiteit Amsterdam, Netherlands \\
    \texttwemoji{sailboat} \texttt{firstname.secondname@cl.cam.ac.uk} \\
    \texttwemoji{herb} \texttt{l.beinborn@vu.nl}}

\begin{document}
\maketitle
\begin{abstract}
We describe our team's contribution to the \textsc{strict-small} track of the BabyLM Challenge \cite{warstadt-et-al-2023-babylm}. The challenge requires training a language model from scratch using only a relatively small training dataset of ten million words. 
We experiment with three variants of cognitively-motivated curriculum learning and analyze their effect on the performance of the model on linguistic evaluation tasks. In the \textbf{vocabulary curriculum}, we analyze methods for constraining the vocabulary in the early stages of training to simulate cognitively more plausible learning curves. In the \textbf{data curriculum} experiments, we vary the order of the training instances based on i) infant-inspired expectations and ii) the learning behaviour of the model. In the \textbf{objective curriculum}, we explore different variations of combining the conventional masked language modelling task with a more coarse-grained word class prediction task to reinforce linguistic generalization capabilities. Our results did not yield consistent improvements over our own non-curriculum learning baseline across a range of linguistic benchmarks; however, we do find marginal gains on select tasks. Our analysis highlights key takeaways for specific combinations of tasks and settings which benefit from our proposed curricula. We moreover determine that careful selection of model architecture, and training hyper-parameters yield substantial improvements over the default baselines provided by the BabyLM challenge. Our code is publicly available at \url{https://github.com/codebyzeb/CLIMB}.
\end{abstract}

\section{Introduction}
Children acquire language skills from being exposed to an estimated two to seven million words per year \citep{gilkerson-et-al-2017}. The current learning regimes of large language models require disproportionately larger sizes of training data to acquire linguistic generalization capabilities \citep{zhang-etal-2021-need}. State-of-the-art LMs are typically trained on gigabytes of data gleaned from the World Wide Web, on multiple GPUs continuously for days at a time \cite{zhao2023survey}. For example, the Chinchilla language model was trained on a dataset of 1.4 trillion words \cite{chinchilla}. Such large-scale training regimes are economically and ecologically unsustainable, and access to the required computing resources remains out of reach for most academic groups and industry start-ups \cite{izsak-etal-2021-train}.

To enable language models to still perform well with limited data, recent work has looked at utilizing smaller, well-curated, and representative corpora \citep{samuel-etal-2023-trained, gao2020pile} and careful selection of training and model hyper-parameters \citep{geiping2023cramming}. `Zero-shot' and `few-shot' learning are other data-efficient approaches which can perform well in certain settings but rely on large pre-trained language models \cite{brown-gpt3,wei-et-al-2021}.
These approaches, however, provide engineering solutions to the problem rather than a cognitively-inspired, compute-efficient framework for training language models from scratch.  

Conventional pre-training of large language models remains far removed from human language learning: models operate on a predetermined static vocabulary and optimize a monotonous training objective on a randomly shuffled dataset. We conducted experiments to explore more dynamic learning processes that are motivated by the psycholinguistic and language acquisition literature and are set within the machine learning paradigm of curriculum learning \cite{bengio2009curriculum}.
Our models are implemented and evaluated within the `BabyLM Challenge' framework, a shared task in which the stated goal is ``to incentivize researchers with an interest in pretraining and/or cognitive modeling to focus their efforts on optimizing pretraining given data limitations inspired by human development'' \cite{warstadt-et-al-2023-babylm}. Our goal in participating in the BabyLM Challenge is two fold: First, we aim to contribute toward democratizing language modelling research and move towards this goal by training smaller language models that are still well-performing on NLP tasks. Second, we establish a computational framework based on curriculum learning for simulating aspects of human language acquisition. We participate in the strictest track of the challenge, limiting the training data to only 10 million words of text extracted from various pre-existing corpora. 

Initially, we train our own BabyBERTa-style vanilla model \footnote{We refer to our non-curriculum learning baselines as `vanilla' models in order to differentiate these models from the baselines that were provided by the workshop organizers.} \cite{huebner-etal-2021-babyberta} and find that simply tuning model size and vocabulary size in itself leads to substantial performance gains on some of the BabyLM test sets compared to the shared task baselines.
We furthermore carried out a number of pre-processing steps on the training data to further improve performance, including concatenating input sequences to make the most of the available input length.

In our own approach, which we term CLIMB \texttwemoji{climbing} -- Curriculum Learning for Infant-inspired Model Building -- we explore three different curriculum strategies for language modelling: gradually increasing the size of the vocabulary (\textbf{vocabulary curriculum}), the difficulty of the training instances (\textbf{data curriculum}), or the specificity of the objective function (\textbf{objective curriculum}) over the course of training. Within the context of the BabyLM Challenge, Curriculum Learning establishes a framework through which we attempt to replicate key facets of child language acquisition. Counter-intuitively, we find that all of our curriculum learning approaches under-perform our BabyBERTa-style (non curriculum learning) vanilla models. Our contribution to the Baby LM Challenge builds upon this negative finding in three main ways:
\begin{enumerate}
    \item Our paper establishes a novel framework through which to categorize and implement curriculum learning methods that simulate human language acquisition. We open-source our accompanying code-base for future research to study how curriculum learning replicates the language learning dynamics in humans. 
    \item We conduct a comprehensive evaluation of our three main curriculum approaches; our results show that the curriculum learning settings we tested did not provide consistent improvements over a baseline on linguistic benchmarks. Instead, we provide a set of recommendations for specific combinations of tasks and settings which may benefit from our proposed curricula. 
    \item We highlight the importance of careful data, model and hyper-parameter selection to establish a well performing fully supervised baseline for the BabyLM shared task. Our vanilla models outperform the shared task baseline models on tasks involving grammatical knowledge (BLiMP: The Benchmark of Linguistic Minimal Pairs \cite{warstadt2020blimp}) and all the shared-task baselines except RoBERTa \cite{liu2019roberta} on tasks involving natural language understanding (SuperGLUE \cite{wang2019superglue}).
\end{enumerate}

\section{Curriculum Learning}
Curriculum learning \cite{bengio2009curriculum} is a machine-learning paradigm which optimizes a model's performance by gradually increasing the difficulty of training over time according to a set schedule (a `curriculum') -- based on the idea that learning should proceed from easy to hard, inspired by the way that humans learn \cite{ELMAN199371}.
Within the context of curriculum learning, one of the central questions is how to define and manipulate the difficulty of the learning process over the course of training. In a recent survey, \citet{soviany2022curriculum} decompose this challenge into two main sub-problems: determining a sorting mechanism to assess the difficulty of instances and developing a pacing function for increasing difficulty over time.

\subsection{Determining Difficulty}
Previous work in curriculum learning typically focuses on difficulty from a data-centric perspective, however, we note that difficulty can arise from (at least) three major elements of training a neural model: the input representation, the data sampling, and the training process. We explore curriculum learning strategies across three distinct dimensions: the vocabulary, the order of training data, and the objective function.

For machine learning models, instance difficulty is in part influenced by the choice of instance representation. For language models, the representational space is constrained by the vocabulary. We propose a new \textbf{vocabulary curriculum} inspired by \citet{soviany2022curriculum}, who discuss linking the curriculum criteria to the observed vocabulary sizes in child development. To the best of our knowledge, this is the first attempt at manipulating the vocabulary available to a language model through curriculum learning.

In natural language processing models, the order of the training instances can have a strong effect on performance \citep{schluter-varab-2018-data}. 
Existing approaches to instance-level curriculum learning determine the difficulty of each instance according to a pre-defined static difficulty assessment according to linguistic criteria \citep{campos2021curriculum,kocmi2017curriculum,liu2018curriculum,platanios2019competence}.
It has been shown that humans pay more attention to stimuli that are in just the right zone of difficulty for them: neither too easy nor too hard \cite{goldilocks}. This so-called `Goldilocks effect' can be modelled by assessing the difficulty of an instance dynamically based on model behaviour \citep{sachan-xing-2016-easy,lalor-yu-2020-dynamic}. Static and dynamic difficulty assessment can be mapped to teacher-centric and learner-centric educational approaches and we compare both variants in our \textbf{data curriculum} experiments. 

Human language learning is guided and enabled to some extent by other agents in the learner's environment (e.g.,\ adult caregivers, siblings) who interact with the learner. In machine learning, such interactions are modelled by the objective function that guides the weight optimization process. The typical `masked language modelling' (MLM) objective function requires that a model predicts a target token from a pre-defined vocabulary of size $N$ given the surrounding context. Thus standard MLM defines an $N$-way token classification task.

Curriculum learning can be leveraged within this context to attenuate the difficulty of the classification task during training. One natural starting point for doing so is to redefine the classification task to be over a smaller set of items, $K$, such that $ K << N$.
\citet{bai-etal-2022-better} map rare words with hypernyms of that word to simplify the classification task in training. A related line of research suggests replacing certain words with either part-of-speech tags \cite{wang2022language} or syntactic dependency relations \cite{cui2022lert}. Since the number of syntactic tags is substantially smaller than the number of vocabulary items, these approaches greatly reduce the difficulty of the objective. Moreover, by varying the amount of syntactic tags that the model should classify over, the difficulty of the task can be dynamically adapted \cite{wang2022language}. We take inspiration from this latter line of work in defining our own \textbf{objective curriculum}.

\subsection{Pacing Functions} Once a notion of difficulty is set, a pacing function is needed to govern how quickly the model will progress from training on easier examples to training on harder ones \cite{wu2021when}. We experiment with two different pacing functions: linear and logarithmic. Linear pacing functions involve a steady and consistent advancement through the curriculum. This approach ensures a gradual increase in difficulty over time. Logarithmic pacing functions, on the other hand, emphasize early exposure to ``easier'' concepts, with diminishing increments as the model's capabilities are assumed to increase. Both pacing functions have been proposed in the broader curriculum learning literature \citep{bai-etal-2022-better, li2021curriculum, wu2021when}.

\begin{table*}
\vspace{-10mm}
\centering
\small
\begin{tabular}{lll}
\toprule
\textbf{Curriculum Type} & \textbf{Parameter} &\textbf{Variants} \\
\midrule
 \multirow{2}{*}{Vocabulary} & Selection & frequency, word class, mixed \\
 & Pacing & linear, logarithmic \\
 \midrule
 \multirow{3}{*}{Data} & Difficulty & source, unigram perplexity, self-perplexity \\
 & Pacing & linear, logarithmic \\
 & Initial Perplexity & unigram, random \\
  \midrule
 \multirow{2}{*}{Objective} & Tasks & noun-verb prediction, POS prediction, MLM\\
 & Learning Setup & sequential, multitask \\
\bottomrule
\end{tabular}
\caption{\label{tbl:configurations} Curriculum learning experiments overview}
\end{table*}

\section{Methodology}

All of our models are based on an 8-layer Transformer language model (Section \ref{subsec:baseline}) comparable to the BabyBERTa model \cite{huebner-etal-2021-babyberta}. 
For all experiments, we use the Hugging Face Transformers library \cite{wolf-etal-2020-transformers}, Weights \& Biases for performance tracking \cite{wandb}, Hydra to define experiment configurations \cite{hydra}, and a high performance computing cluster.

We introduce curriculum learning to three of the primary components of language model pre-training: the vocabulary (Section \ref{subsec:vocab-cl}), the data sampling approach (Section \ref{subsec:data-cl}), and the selection of the objective function (Section \ref{subsec:objective-cl}). For each of these aspects, we attempt to simulate facets of human language learning by dynamically increasing the difficulty of the language modelling task over the course of training. Table~\ref{tbl:configurations} provides an overview of our experiment variables.

\subsection{Training Data}
\label{subsec:data}
We use only the training data provided in the \textsc{strict-small} track of the BabyLM challenge, which is limited to 10 million words and combined from 10 individual corpora. Given the variety of data sources (including books, subtitles, transcripts and articles) we carefully curated the data to ensure consistency across corpora. These steps include lowercasing, normalizing punctuation, standardizing typographical conventions using regular expressions, and removing extraneous lines (such as page numbers, bibliography entries, plain text tables , and one-word on-screen actions). We also concatenated contiguous sections of five lines into a single data instance in the transcribed speech corpora (except the BNC) due to the relatively short sequence lengths. In addition, we join data at the point of passing input to the models, in order to make full use of the available input sequence length (128 subtokens).
 
According to the rules of the \textsc{strict-small} track, we were not permitted to make use of external resources, including supervised part-of-speech (POS) taggers. Therefore, we attempted to cluster the words in the training data into word classes by applying the \texttt{anchor-features} algorithm of the unsupervised POS-tagger by \citet{stratos-etal-2016-unsupervised} on our cleaned data. The algorithm yields 30 clusters which we manually mapped to the 12 universal speech tags \citep{petrov-etal-2012-universal} by choosing the POS-tag that best represents the anchor word of each cluster. We were only able to identify 10 of the 12 universal POS tags in the 30 clusters: no cluster neatly coincided with 'ADV' or 'X' tags. We provide further detail on our data pre-processing and unsupervised POS-tagging in the Appendix.

We provide our cleaned and tagged versions of the 10M word dataset on Hugging Face, along with the scripts used.\footnote{\url{https://huggingface.co/cambridge-climb}} Our pre-processing procedure reduces the data down to 335,858 instances (corresponding to roughly 9.4 million words) from the initial 1,058,740 newline-delineated samples.\footnote{The word count is estimated by whitespace splitting; the same metric used by the organizers of the task to derive the estimate of 10 million words. When applying a tokenizer, the pre-processed dataset is more accurately split into 11.7 million words (including punctuation) or 13.6 million subwords} Our models, tokenizers and part-of-speech taggers were trained on this pre-processed data; however, we actually noticed an increase in performance when training on the raw data, as discussed in Section~\ref{sec:discussion}.

\subsection{Vanilla Models}
\label{subsec:baseline}

\begin{table*}
\centering
\small
\begin{tabular}{l | rrrrr | rrrr}
\toprule
Model  & Layers & Heads & Hidden & Intermediate & Vocab & Train.steps & BLiMP & BLiMP.Supp & Perplexity \\
\midrule
Small  & 8 & 8 & 256 & 2,048   & 8,192   & 250K      & 75.43      & 61.14       & 9.46    \\

Medium & 10 & 10 & 500 & 2,000 & 8,192  & 156K      & 76.45      & 63.28        & 9.05  \\
Large  & 12 & 12 & 768 & 3,072 & 8,192   & 94K      & 75.80      & 60.83      & 9.34 \\[2mm]
\hline \\
Small  & 8 & 8 & 256 & 2,048   & 16,384  & 250K      & 76.16      & 60.85       & 13.80    \\
Medium & 10 & 10 & 500 & 2,000  & 16,384 & 94K      & 76.09      & 60.03        & 13.80     \\
Large  & 12 & 12 & 768 & 3,072 & 16,384  & 62K      & 75.08      & 63.45      & 14.22     \\
\bottomrule
\end{tabular}
\caption{\label{tbl:baseline-size-comparison} Our vanilla BabyBERTa-style models evaluated on original BLiMP and the BLiMP-like tasks prepared for BabyLM (BLiMP.Supp). Models are grouped by their vocabulary sizes.}
\end{table*}

We investigate three different sizes of a vanilla Pre-Layer Norm RoBERTa model \cite{liu2019roberta,Ott2019fairseqAF} based on the BabyBERTa model \cite{huebner-etal-2021-babyberta}: `small', `medium', and `large' -- Table \ref{tbl:baseline-size-comparison} lists the model configurations and presents the results for the different model sizes evaluated by perplexity, on BLiMP \cite{warstadt2020blimp} and on the supplementary BLiMP-like tasks issued by the BabyLM organizers (`Blimp.Supp'). We found the medium model with a small vocabulary size performed the best overall; however, the small model achieved similar results, and so to save on compute and keep to the restrained intentions of the \textsc{strict-small} track, we used the small model in our curriculum learning experiments.
We use Byte Pair Encoding (BPE) tokenization \cite{gage1994new} with a vocabulary of 8,192 because it yields better overall performance compared to a larger vocabulary of 16,384. The tokenizers we use in our experiments were trained on the cleaned data that we processed using the steps outlined in \ref{subsec:data}. In pilot experiments, we did not observe the benefits reported by \citet{huebner-etal-2021-babyberta} from removing the unmasking procedure that is a standard component of the MLM objective \cite{devlin-etal-2019-bert}, and therefore did not investigate this option further.

All of the curriculum learning methods in the following sections were applied on top of our small vanilla BabyBERTa-style baseline -- to isolate the effect of the curriculum-learning training process, we fixed the architecture of the model and the model hyper-parameters. We use an AdamW optimizer with linear scheduling \cite{loshchilov2019decoupled}.

\subsection{Vocabulary Curriculum}
\label{subsec:vocab-cl}

During the early stages of language acquisition, children start with a small vocabulary that rapidly expands at a rate of eight to ten words per day \cite{weizman2001lexical}. In this process, children prioritize learning verbs and nouns before progressing to other parts of speech \cite{bergelson2015early}. Large language models, on the other hand, tend to begin training with a full, fixed vocabulary available to them. 

To represent a child's growing vocabulary, we select a limited vocabulary in the initial stages of learning and map all other input tokens into the representation for the unknown token (\textsc{UNK}). We consider three strategies for selecting tokens. In the first strategy, tokens are selected according to frequency. We approximate the frequency of a token by the identifier the BPE tokenizer assigns to it as lower IDs are assigned to tokens that are merged first (i.e., sequences of characters that occur more frequently in the corpus). In the second strategy, tokens are selected by their word class. We approximate the word class of a token by the cluster that the unsupervised POS-tagger assigns to it. We order the word classes as follows, progressing from lexical to functional classes per \citet{bergelson2015early}: NOUN, VERB, ADJ, PRON, DET, ADP, NUM, CONJ, PRT, PNCT. In this strategy, all words with the respective part-of-speech tag are included in the vocabulary at the same step during learning. To smooth this process, we combine the frequency and the word class constraint in the third strategy.
We sort words by their frequency (approximated by the token ID) within each part-of-speech category. Note that the same word may be available in some instances and not others if it is assigned a more difficult POS tag. 

During the initial steps of training, only 10\% of the tokens are available while the rest are replaced with UNK. The vocabulary curriculum regime begins after 25,000 training steps and ends at 350,000 steps, during which time, the vocabulary gradually increases according to a pacing function. We experiment with linear and logarithmic pacing functions.  After the end of the curriculum regime, there remain 50,000 training steps before the end of training during which all of the vocabulary tokens are available to the model. Figure \ref{fig:pacing_fn} in the Appendix shows a plot of the percentage of unmasked vocabulary over the course of training according to our pacing functions.

\subsection{Data Curriculum}
\label{subsec:data-cl}

Conventional masked language modelling approaches train a given neural network on a large amount of crawled internet data. The resulting text sequences are usually not curated beyond basic cleaning and are presented to the model in random order, in contrast to the way that human children learn a language.
 
We attempt to carefully optimize the way data is sampled and presented to the language model over the course of training. We experiment with theory-driven and model-driven approaches to determine the `relative difficulty' of a certain example and train the model on instances with progressively increasing difficulty.

\paragraph{Source Difficulty} We order the available datasets based on their sources so that spoken samples are considered `easier' and purely written texts `harder', following the findings of \citet{huebner-etal-2021-babyberta}. Within this ordering, we place the mostly child-directed speech from CHILDES before adult-to-adult dialogues in the Switchboard Corpus, and Simple Wikipedia before Wikipedia, see Table \ref{tbl:source_order}.\footnote{There is likely some adult-to-adult dialogue included in CHILDES as well.}

\begin{table}
\centering
\small
\begin{tabular}{cl}
\toprule
Difficulty Level & Corpora \\
\midrule
1 & AO-CHILDES\\
2 & BNC Spoken, Switchboard \\
3 & Open Subtitles, QED \\
4 & CBT, Children's Stories \\
5 & Simple Wikipedia \\
6 & Wikipedia, Gutenberg \\
\bottomrule
\end{tabular}
\caption{\label{tbl:source_order} Difficulty level assigned to each dataset.}
\end{table}

\paragraph{Model Difficulty}
Determining the difficulty of an instance based on its data source is a relatively naive heuristic that ignores the variation of instance difficulty within one corpus. As a more fine-grained alternative, we determine the difficulty of each instance individually using the model-intrinsic metric of perplexity which determines the likelihood of a sentence. We experiment with two variants: a static unigram language model and a more dynamic self-evaluation. With the unigram model, perplexity for each instance is only determined once at the beginning of training. Alternatively, we evaluate the perplexity of the remaining training data using the model that has been trained so far -- from model checkpoints saved at regular intervals in training (every 25K steps). 

One challenge with the latter approach is the lack of exposure to training data at the beginning, leading to random perplexity scores for each sample. To address this, we propose two ideas: 1) using a separately trained unigram model to initially evaluate perplexity, or 2) initially sample training instances randomly. After 25,000 training steps, we switch to using the current model for perplexity evaluation. Every 25,000 steps thereafter, we re-evaluate perplexity to identify samples categorized as relatively difficult or relatively easy by the model.

\subsection{Objective Curriculum}
\label{subsec:objective-cl}

The MLM objective has proven tremendously successful in training Transformer networks as language models \cite{devlin-etal-2019-bert}. Psycholinguistic research, however, suggests that MLM is not a cognitively plausible approximation of language acquisition processes in children \cite{caucheteux2023evidence}.
Curriculum learning establishes a framework for varying the difficulty of the learning process over the course of training. The MLM objective is a very challenging discriminative classification task because the identity of the masked token needs to be determined over the entire vocabulary. We experiment with using more coarse-grained tasks at the initial stages of training to facilitate generalization and leverage syntactic information. Research in cognitive linguistics has shown that one-year-old infants are sensitive to distributional aspects of language and from two years of age begin to recognize lexical categories such as nouns and verbs \citet{alishahi2010computational, gleitman1990structural}. We therefore experiment with predicting only the word class of a masked token at the start of training rather than predicting its exact target token ID. 

The psycholinguistic literature remains divided on the question of how exactly word learning proceeds from memorizing a small set of fixed lexical items to a more generalized representation of word classes \cite{clark2015first}. Our framework provides a flexible approach to vary the difficulty of objective functions during the course of training, and to enable systematic studies of the effect of objective functions on the acquisition of linguistic knowledge by a model.
Here we propose estimating the word class using the unsupervised POS tagger and we vary the number of POS tags which are being classified over. The masked word is classified into 1) one of VERB, NOUN, or OTHER, or 2) one of 10 universal POS tags. 

We examine activating the tasks in sequential order (first word class prediction then MLM) or optimizing them in parallel, comparable to a multi-task learning setting. For each objective function, we learn a separate task head with its own linear task classifier and separate optimizer.

\begin{figure*}
\centering
\includegraphics[height=9.5cm]{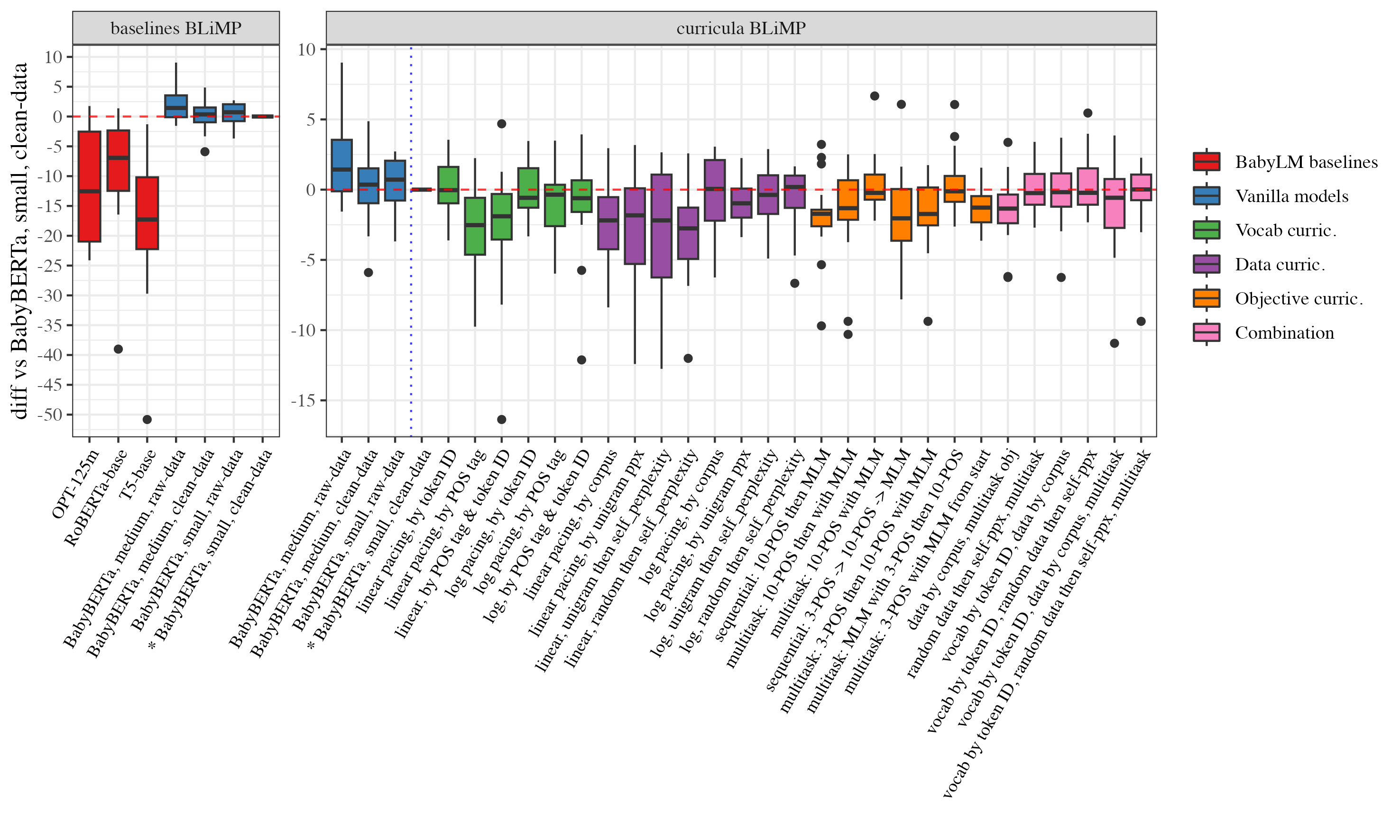}
\caption{\label{fig:blimp-boxplots} Comparison of the BabyLM baselines with our BabyBERTa-style vanilla models (left), and our vanilla models against our curriculum learning models (right) -- using BabyBERTa-small trained on clean data as a reference point (asterisked) to show the difference in scores on BLiMP and BLiMP-supplement tasks. For combination models, all pacing is logarithmic, and `multitask' refers to the 2-task objective curriculum, 10 POS-tags and MLM from the outset. Absolute values may be found in Appendix Tables~\ref{tbl:result-vocab-cl}--\ref{tbl:submission-comparison}.
}
\end{figure*}

\section{Results}

Multiple evaluation metrics are employed in BabyLM. In this paper we focus on BLiMP \cite{warstadt2020blimp} and the supplementary BLiMP-style tests provided by the shared task organizers. We also report our results on the natural language understanding benchmark, SuperGLUE \cite{wang2019superglue}, and the ambiguous subset of MSGS (the Mixed Signals Generalization Set) \cite{warstadt-etal-2020-learning}. In brief, BLiMP evaluates specific linguistic abilities, MSGS evaluates linguistic preference over surface generalisation and SuperGLUE evaluates downstream task performance. For all scores, we report the average score across all categories, rather than test instances, as provided by the BabyLM evaluation pipeline.\footnote{For instance, there are 12 categories in BLiMP but 50+ individual tests. We average over the scores given for each category, rather than the scores given for each test.} All of our curriculum learning models are small BabyBERTa-style ones using the parameters shown in Table~\ref{tbl:baseline-size-comparison} and the cleaned training dataset of 9.4M words (reduced from the 10M word dataset for the \textsc{strict-small} track) and their results can be found in Tables \ref{tbl:result-vocab-cl}, \ref{tbl:result-data-cl} and \ref{tbl:result-obj-cl}. 

In the tables we compare to our small BabyBERTa-style vanilla model also trained on the clean data (Section \ref{subsec:baseline}). Figure~\ref{fig:blimp-boxplots} visualizes these comparisons for the BLiMP tasks; there are  similar plots for SuperGLUE in the Appendix (Figure~\ref{fig:glue-boxplots}).
Furthermore, we experimented with some combinations of different curricula to see how they would interact (Table~\ref{tbl:result-combination-cl}), and compare the official BabyLM shared-task baselines with our shared task entries -- a number of our own BabyBERTa-style vanilla models and curriculum learning models (Table~\ref{tbl:submission-comparison}). For all of our runs, we use the same set of hyper-parameters that we report in Table~\ref{tbl:baseline_hyperparams}. We also report the average amount of compute used for each type of curriculum learning setting (Table~\ref{tbl:compute}).

We find notable gains for our own vanilla models over the shared-task baselines, and, while we do not identify further large improvements in our curriculum learning models, we do notice some modest gains which suggest possibilities for future research and experimentation over variables. While the differences in performance between most of our experimental conditions are small, the large number of ablations we run enables us to provide a comprehensive set of recommendations for how and when different curriculum learning strategies may offer improved performance on linguistic tasks.
Below we summarize our observations over the full results tables.

\paragraph{In general, log pacing works at least as well as linear pacing across different curricula learning strategies.}
In our data curriculum experiments, models using the log pacing function outperform their linear counterparts in 4/4 settings on BLiMP, and 3/4 settings for BLiMP-supplement and SuperGLUE (Table~\ref{tbl:result-data-cl}). This indicates that rapidly increasing the difficulty of training instances in the early stages brings downstream benefits on grammaticality and NLU tasks.

In our vocabulary curriculum experiments on the other hand, there is not such a clear picture. Log pacing outperforms linear in 2/3 settings on BLiMP and 3/3 on SuperGLUE, but 0/3 for BLiMP-supplement (Table~\ref{tbl:result-vocab-cl}).
Presumably this is a reflection of the different vocabulary required by each set of evaluation tasks, which could be a matter for future investigation but also indicates that we do not yet have a clear generalizable pacing function for the vocabulary curriculum. There are of course other pacing functions to be tried.

\paragraph{Different representations of vocabulary difficulty work better for different tasks.}
When representing difficulty in the vocabulary curriculum experiments, token ID -- our proxy for frequency -- appears to work better than word classes (POS tags) or a combination of token ID and POS tags on the BLiMP evaluation tasks, but worse than POS tags on SuperGLUE and MSGS (Table~\ref{tbl:result-vocab-cl}).

\paragraph{In multi-corpora datasets, ordering by difficulty is a good first step.}
Training data requirements have grown so much in modern NLP that usually training a language model from scratch will involve multiple datasets, or multiple domains. The results of our data curriculum experiments indicate that a good first step is to put these sub-corpora into some order of intuitive difficulty, as we did (Table~\ref{tbl:result-data-cl}). In the case of BLiMP this approach outperforms our perplexity-based data curricula, and with log pacing our vanilla model. The same is true of MSGS  (with log pacing), as well as BLiMP-supplement and SuperGLUE (though the last two do not beat our vanilla model). 
Amongst the perplexity-driven models, the picture is less positive: out of 24 tests, only one model outperforms our vanilla model (log pacing, random initialisation + model perplexity in Table~\ref{tbl:result-data-cl}).

\paragraph{Multitask learning holds sway over sequentially swapping objective functions for now.}
In our experiments with curricula for the objective function, we compare training on simultaneous tasks -- known as multitask learning \cite{caruana1997multitask} -- with predefined sequences of objective functions which swap from one to another at set thresholds in the training process. We set up two sequential curricula: one with 2 tasks (predicting the 10 universal POS tags found in our dataset, and MLM) and the other with 3 (like the 2 task curriculum, additionally with noun/verb/other prediction). We compare these against multitasking alternatives. In general the sequential curricula are outperformed by the multitasking ones, though the 3-task sequential curriculum outperforms our BabyBERTa-style vanilla model on SuperGLUE and is second only marginally to our best-performing multitask model (Table~\ref{tbl:result-obj-cl}). The multitask learning model with 10-class universal POS-tag prediction and MLM in place from the outset performs best on BLiMP and SuperGLUE. However, our best model on BLiMP-supplement -- a multitask one -- has an element of sequential task scheduling in that the two POS-tag prediction tasks are lined up one after the other, with a switch from 3-class to 10-class after 6.25\% of training steps. In Figure \ref{fig:baseline_obj_cl_blimp_supp}, we visualize this result for each task in BLiMP-supplement, illustrating that our curriculum learning model improves over our vanilla model in 5/6 tasks.
Altogether, these results suggest that sequential objective function curricula do hold some potential for performance gains if further tuning of the tasks and scheduling can be carried out.

\paragraph{Combining all three curricula shows potential on BLiMP.}
While each individual curriculum learning experiment did not result in consistent improvements across tasks, we investigated whether combining aspects from the different curricula would, together, improve the model.
We do find that a combination of all three curricula outperforms any single curriculum model on BLiMP, but the same is not true for BLiMP-supplement and SuperGLUE (Table~\ref{tbl:result-combination-cl}). This is another matter for future investigation, as it seems that improving each of the three curricula we investigate may lead to further gains if they are all combined.

\paragraph{In small data settings, filtering data which we intuitively think is noisy is in fact counter-productive.} Perhaps surprisingly, we find that the vanilla models trained on the raw data outperform those trained on the pre-processed data on BLiMP and MSGS. We surmise that models can learn even from linguistically non-standard datapoints.

\subsection{Submitted models}

Table \ref{tbl:submission-comparison} in the Appendix compares our submissions to the shared task baselines. We submitted our best curriculum learning models from each individual curriculum learning setting, and four different vanilla models: two small and two medium models, where each pair additionally varies by whether it was trained on the pre-processed dataset or the raw dataset.
We find our curriculum learning models are comparable to our BabyBERTa-style vanilla models, and we think that in most cases some continued experimentation with configurations may yield larger gains for CL approaches.

For interest, we also trained a BabyBERTa-style large vanilla model on the 100M training set made available in the BabyLM \textsc{strict} track (`large-100M' in the table). The improvements over smaller models trained on less data are evident and finally provide an advantage over the RoBERTa baseline on SuperGLUE. It remains to be seen how well curriculum learning methods, and our preprocessing methods, would work with this larger dataset.

\vspace{-1mm}

\section{Discussion}\label{sec:discussion}

We set out to investigate a number of curriculum learning approaches to language model training, motivated by findings from the human language acquisition process and by the wish to successfully train smaller models for smaller budgets.
We first of all implemented a stronger model of our own, based on BabyBERTa \cite{huebner-etal-2021-babyberta} and found that a small 8-layer vanilla model could outperform the provided BabyLM baselines on the BLiMP grammaticality tests and get close to the best RoBERTa shared-task baseline on SuperGLUE. This underlines the findings reported in the BabyBERTa paper: that with smaller datasets, it makes sense to use smaller models and a smaller vocabulary size.

The results of our curriculum learning experiments, trained with a small BabyBERTa-style vanilla model, suggest that we can further improve performance in certain linguistic tasks by careful application of a pacing function, how we represent and grow the model's vocabulary during training, select the next training instances according to their difficulty, and vary the objective function. Specifically, we find that a logarithmic pacing function works better for the data curriculum than a linear one, but the findings for the vocabulary curriculum are less clear. Other pacing functions might be tried in the future, including those that reflect acquisition theory around non-monotonic or `U-shaped' development trajectories.

It is apparent that ordering the subcorpora within a training set may be worthwhile, and that perplexity-based approaches to data selection hold potential even though we have not found a clear-cut best method for perplexity calculation as yet. As shown in other NLP work, multitask learning can be a beneficial approach, though MLM or next-word prediction remain preeminent as singular tasks used in language modelling. We find multitask learning models hard to beat in the objective curriculum, but do find good performance in our sequential settings. We believe that future work varying the timing of task switches and introducing more tasks could be worthwhile.

On a more general note, the Baby LM challenge evaluates a language model only on its final downstream performance on a set of tasks -- i.e.\ at a finite point in time. The challenge does not directly measure whether a given model is learning in a `human-like' fashion. Our contribution to the BabyLM challenge is to provide a set of curriculum learning strategies which are motivated by the language learning dynamics of infants and children. We encourage future research to study how to quantitatively evaluate whether the learning trajectory of a model parallels that of a human language learner and how similarities to human language learning results in downstream NLU performance.

\vspace{-1mm}

\section{Conclusions}
We use child-like language learning as inspiration to investigate and implement three types of curriculum learning for language modelling: gradually increasing the size of the vocabulary (\textbf{vocabulary curriculum}), the difficulty of the training instances (\textbf{data curriculum}), or the specificity of the objective function (\textbf{objective curriculum}).

We find that our BabyBERTa-style vanilla models outperform the BabyLM baselines on BLiMP and MSGS, and get close on SuperGLUE. Our various curriculum learning models at times offer further gains over our vanilla models, and indicate the potential for curriculum learning methods given further exploration. We list out a set of recommendations for when and how to optimally apply our proposed curriculum learning strategies.

Additionally, training our vanilla model trained on unprocessed data outperforms a `cleaned' version -- suggesting that retaining as much data as possible, in low-resource settings, is more important than standardizing it according to linguistic norms.

Finally, our work establishes a computational framework for how to categorise and implement curricula learning strategies that simulate human language learning dynamics. 

\section*{Acknowledgements}

This paper reports on work supported by Cambridge University Press \& Assessment.
It was performed using resources provided by the Cambridge Service for Data Driven Discovery (CSD3) operated by the University of Cambridge Research Computing Service, provided by Dell EMC and Intel using Tier-2 funding from the Engineering and Physical Sciences Research Council (capital grant EP/T022159/1), and DiRAC funding from the Science and Technology Facilities Council. Richard Diehl Martinez is supported by the Gates Cambridge Trust (grant OPP1144 from the Bill \& Melinda Gates Foundation).
Hope McGovern's work is supported by The Cambridge Trust and the Woolf Institute for Interfaith Relations.
Z\'ebulon Goriely's work is supported by The Cambridge Trust.
Lisa Beinborn's work is supported by the Dutch National Science Organisation (NWO) through the VENI program (Vl.Veni.211C.039).

\bibliography{anthology,bib}

\begin{thebibliography}{49}
\expandafter\ifx\csname natexlab\endcsname\relax\def\natexlab#1{#1}\fi

\bibitem[{Alishahi(2010)}]{alishahi2010computational}
Afra Alishahi. 2010.
\newblock \emph{Computational modeling of human language acquisition}.
\newblock Morgan \& Claypool Publishers.

\bibitem[{Bai et~al.(2022)Bai, Wang, Sordoni, and Shi}]{bai-etal-2022-better}
He~Bai, Tong Wang, Alessandro Sordoni, and Peng Shi. 2022.
\newblock \href {https://doi.org/10.18653/v1/2022.acl-long.96} {Better language
  model with hypernym class prediction}.
\newblock In \emph{Proceedings of the 60th Annual Meeting of the Association
  for Computational Linguistics (Volume 1: Long Papers)}, pages 1352--1362,
  Dublin, Ireland. Association for Computational Linguistics.

\bibitem[{Bengio et~al.(2009)Bengio, Louradour, Collobert, and
  Weston}]{bengio2009curriculum}
Yoshua Bengio, J{\'e}r{\^o}me Louradour, Ronan Collobert, and Jason Weston.
  2009.
\newblock Curriculum learning.
\newblock In \emph{Proceedings of the 26th Annual International Conference on
  Machine Learning}, pages 41--48.

\bibitem[{Bergelson and Swingley(2015)}]{bergelson2015early}
Elika Bergelson and Daniel Swingley. 2015.
\newblock Early word comprehension in infants: Replication and extension.
\newblock \emph{Language Learning and Development}, 11(4):369--380.

\bibitem[{Biewald(2020)}]{wandb}
Lukas Biewald. 2020.
\newblock \href {https://www.wandb.com/} {Experiment tracking with {W}eights
  and {B}iases}.

\bibitem[{Bird et~al.(2009)Bird, Klein, and Loper}]{bird2009natural}
Steven Bird, Ewan Klein, and Edward Loper. 2009.
\newblock \emph{Natural language processing with Python: analyzing text with
  the natural language toolkit}.
\newblock " O'Reilly Media, Inc.".

\bibitem[{Brown et~al.(2020)Brown, Mann, Ryder, Subbiah, Kaplan, Dhariwal,
  Neelakantan, Shyam, Sastry, Askell, Agarwal, Herbert-Voss, Krueger, Henighan,
  Child, Ramesh, Ziegler, Wu, Winter, Hesse, Chen, Sigler, Litwin, Gray, Chess,
  Clark, Berner, McCandlish, Radford, Sutskever, and Amodei}]{brown-gpt3}
Tom Brown, Benjamin Mann, Nick Ryder, Melanie Subbiah, Jared~D Kaplan, Prafulla
  Dhariwal, Arvind Neelakantan, Pranav Shyam, Girish Sastry, Amanda Askell,
  Sandhini Agarwal, Ariel Herbert-Voss, Gretchen Krueger, Tom Henighan, Rewon
  Child, Aditya Ramesh, Daniel Ziegler, Jeffrey Wu, Clemens Winter, Chris
  Hesse, Mark Chen, Eric Sigler, Mateusz Litwin, Scott Gray, Benjamin Chess,
  Jack Clark, Christopher Berner, Sam McCandlish, Alec Radford, Ilya Sutskever,
  and Dario Amodei. 2020.
\newblock \href
  {https://proceedings.neurips.cc/paper_files/paper/2020/file/1457c0d6bfcb4967418bfb8ac142f64a-Paper.pdf}
  {Language models are few-shot learners}.
\newblock In \emph{Advances in Neural Information Processing Systems},
  volume~33, pages 1877--1901. Curran Associates, Inc.

\bibitem[{Campos(2021)}]{campos2021curriculum}
Daniel Campos. 2021.
\newblock \href {http://arxiv.org/abs/2108.02170} {Curriculum learning for
  language modeling}.

\bibitem[{Caruana(1997)}]{caruana1997multitask}
Rich Caruana. 1997.
\newblock Multitask learning.
\newblock \emph{Machine learning}, 28:41--75.

\bibitem[{Caucheteux et~al.(2023)Caucheteux, Gramfort, and
  King}]{caucheteux2023evidence}
Charlotte Caucheteux, Alexandre Gramfort, and Jean-R{\'e}mi King. 2023.
\newblock Evidence of a predictive coding hierarchy in the human brain
  listening to speech.
\newblock \emph{Nature human behaviour}, 7(3):430--441.

\bibitem[{Clark and Casillas(2015)}]{clark2015first}
Eve~V Clark and Marisa Casillas. 2015.
\newblock \emph{First language acquisition}, pages 167--167. Routledge.

\bibitem[{Cui et~al.(2022)Cui, Che, Wang, and Liu}]{cui2022lert}
Yiming Cui, Wanxiang Che, Shijin Wang, and Ting Liu. 2022.
\newblock \href {https://arxiv.org/abs/2211.05344} {Lert: A
  linguistically-motivated pre-trained language model}.
\newblock arXiv:2211.05344.

\bibitem[{Devlin et~al.(2019)Devlin, Chang, Lee, and
  Toutanova}]{devlin-etal-2019-bert}
Jacob Devlin, Ming-Wei Chang, Kenton Lee, and Kristina Toutanova. 2019.
\newblock \href {https://doi.org/10.18653/v1/N19-1423} {{BERT}: Pre-training of
  deep bidirectional transformers for language understanding}.
\newblock In \emph{Proceedings of the 2019 Conference of the North {A}merican
  Chapter of the Association for Computational Linguistics: Human Language
  Technologies, Volume 1 (Long and Short Papers)}, pages 4171--4186,
  Minneapolis, Minnesota. Association for Computational Linguistics.

\bibitem[{Elman(1993)}]{ELMAN199371}
Jeffrey~L. Elman. 1993.
\newblock \href {https://doi.org/https://doi.org/10.1016/0010-0277(93)90058-4}
  {Learning and development in neural networks: the importance of starting
  small}.
\newblock \emph{Cognition}, 48(1):71--99.

\bibitem[{Gage(1994)}]{gage1994new}
Philip Gage. 1994.
\newblock A new algorithm for data compression.
\newblock \emph{C Users Journal}, 12(2):23--38.

\bibitem[{Gao et~al.(2020)Gao, Biderman, Black, Golding, Hoppe, Foster, Phang,
  He, Thite, Nabeshima, Presser, and Leahy}]{gao2020pile}
Leo Gao, Stella Biderman, Sid Black, Laurence Golding, Travis Hoppe, Charles
  Foster, Jason Phang, Horace He, Anish Thite, Noa Nabeshima, Shawn Presser,
  and Connor Leahy. 2020.
\newblock \href {http://arxiv.org/abs/2101.00027} {The pile: An 800gb dataset
  of diverse text for language modeling}.

\bibitem[{Geiping and Goldstein(2023)}]{geiping2023cramming}
Jonas Geiping and Tom Goldstein. 2023.
\newblock Cramming: Training a language model on a single {GPU} in one day.
\newblock In \emph{International Conference on Machine Learning}, pages
  11117--11143. PMLR.

\bibitem[{Gilkerson et~al.(2017)Gilkerson, Richards, Warren, Montgomery,
  Greenwood, Oller, Hansen, and Paul}]{gilkerson-et-al-2017}
Jill Gilkerson, Jeffrey~A Richards, Steven~F Warren, Judith~K Montgomery,
  Charles~R Greenwood, D~Kimbrough Oller, John H~L Hansen, and Terrance~D Paul.
  2017.
\newblock \href {https://doi.org/https://doi.org/10.1044/2016_AJSLP-15-0169}
  {Mapping the early language environment using all-day recordings and
  automated analysis}.
\newblock \emph{American Journal of Speech-Language Pathology}, 26:248–265.

\bibitem[{Gleitman(1990)}]{gleitman1990structural}
Lila Gleitman. 1990.
\newblock The structural sources of verb meanings.
\newblock \emph{Language acquisition}, 1(1):3--55.

\bibitem[{Hoffmann et~al.(2022)Hoffmann, Borgeaud, Mensch, Buchatskaya, Cai,
  Rutherford, de~Las~Casas, Hendricks, Welbl, Clark, Hennigan, Noland,
  Millican, van~den Driessche, Damoc, Guy, Osindero, Simonyan, Elsen, Rae,
  Vinyals, and Sifre}]{chinchilla}
Jordan Hoffmann, Sebastian Borgeaud, Arthur Mensch, Elena Buchatskaya, Trevor
  Cai, Eliza Rutherford, Diego de~Las~Casas, Lisa~Anne Hendricks, Johannes
  Welbl, Aidan Clark, Tom Hennigan, Eric Noland, Katie Millican, George van~den
  Driessche, Bogdan Damoc, Aurelia Guy, Simon Osindero, Karen Simonyan, Erich
  Elsen, Jack~W. Rae, Oriol Vinyals, and Laurent Sifre. 2022.
\newblock Training compute-optimal large language models.
\newblock In \emph{Proceedings of the 36th Conference on Neural Information
  Processing Systems (NeurIPS)}.

\bibitem[{Huebner et~al.(2021)Huebner, Sulem, Cynthia, and
  Roth}]{huebner-etal-2021-babyberta}
Philip~A. Huebner, Elior Sulem, Fisher Cynthia, and Dan Roth. 2021.
\newblock \href {https://doi.org/10.18653/v1/2021.conll-1.49} {{B}aby{BERT}a:
  Learning more grammar with small-scale child-directed language}.
\newblock In \emph{Proceedings of the 25th Conference on Computational Natural
  Language Learning}, pages 624--646, Online. Association for Computational
  Linguistics.

\bibitem[{Izsak et~al.(2021)Izsak, Berchansky, and
  Levy}]{izsak-etal-2021-train}
Peter Izsak, Moshe Berchansky, and Omer Levy. 2021.
\newblock \href {https://doi.org/10.18653/v1/2021.emnlp-main.831} {How to train
  {BERT} with an academic budget}.
\newblock In \emph{Proceedings of the 2021 Conference on Empirical Methods in
  Natural Language Processing}, pages 10644--10652, Online and Punta Cana,
  Dominican Republic. Association for Computational Linguistics.

\bibitem[{Kidd et~al.(2012)Kidd, Piantadosi, and Aslin}]{goldilocks}
Celeste Kidd, Steven~T. Piantadosi, and Richard~N. Aslin. 2012.
\newblock \href {https://doi.org/10.1371/journal.pone.0036399} {The {Goldilocks
  Effect}: Human infants allocate attention to visual sequences that are
  neither too simple nor too complex}.
\newblock \emph{PLOS ONE}, 7(5):1--8.

\bibitem[{Kocmi and Bojar(2017)}]{kocmi2017curriculum}
Tom Kocmi and Ond{\v{r}}ej Bojar. 2017.
\newblock Curriculum learning and minibatch bucketing in neural machine
  translation.
\newblock In \emph{Proceedings of the International Conference on Recent
  Advances in Natural Language Processing, RANLP 2017}, pages 379--386.

\bibitem[{Lalor and Yu(2020)}]{lalor-yu-2020-dynamic}
John~P. Lalor and Hong Yu. 2020.
\newblock \href {https://doi.org/10.18653/v1/2020.findings-emnlp.48} {Dynamic
  data selection for curriculum learning via ability estimation}.
\newblock In \emph{Findings of the Association for Computational Linguistics:
  EMNLP 2020}, pages 545--555, Online. Association for Computational
  Linguistics.

\bibitem[{Li et~al.(2021)Li, Zhang, and He}]{li2021curriculum}
Conglong Li, Minjia Zhang, and Yuxiong He. 2021.
\newblock \href {http://arxiv.org/abs/2108.06084} {Curriculum learning: {A}
  regularization method for efficient and stable billion-scale {GPT} model
  pre-training}.
\newblock \emph{CoRR}, abs/2108.06084.

\bibitem[{Liu et~al.(2018)Liu, He, Liu, Zhao et~al.}]{liu2018curriculum}
Cao Liu, Shizhu He, Kang Liu, Jun Zhao, et~al. 2018.
\newblock Curriculum learning for natural answer generation.
\newblock In \emph{IJCAI}, pages 4223--4229.

\bibitem[{Liu et~al.(2019)Liu, Ott, Goyal, Du, Joshi, Chen, Levy, Lewis,
  Zettlemoyer, and Stoyanov}]{liu2019roberta}
Yinhan Liu, Myle Ott, Naman Goyal, Jingfei Du, Mandar Joshi, Danqi Chen, Omer
  Levy, Mike Lewis, Luke Zettlemoyer, and Veselin Stoyanov. 2019.
\newblock \href {https://arxiv.org/abs/1907.11692} {{RoBERTa}: A robustly
  optimized {BERT} pretraining approach}.
\newblock arXiv:1907.11692.

\bibitem[{Loshchilov and Hutter(2019)}]{loshchilov2019decoupled}
Ilya Loshchilov and Frank Hutter. 2019.
\newblock \href {https://arxiv.org/abs/1711.05101} {Decoupled weight decay
  regularization}.
\newblock arXiv:1711.05101.

\bibitem[{Ott et~al.(2019)Ott, Edunov, Baevski, Fan, Gross, Ng, Grangier, and
  Auli}]{Ott2019fairseqAF}
Myle Ott, Sergey Edunov, Alexei Baevski, Angela Fan, Sam Gross, Nathan Ng,
  David Grangier, and Michael Auli. 2019.
\newblock \href {https://api.semanticscholar.org/CorpusID:91184134} {fairseq: A
  fast, extensible toolkit for sequence modeling}.
\newblock In \emph{North American Chapter of the Association for Computational
  Linguistics}.

\bibitem[{Petrov et~al.(2012)Petrov, Das, and
  McDonald}]{petrov-etal-2012-universal}
Slav Petrov, Dipanjan Das, and Ryan McDonald. 2012.
\newblock \href
  {http://www.lrec-conf.org/proceedings/lrec2012/pdf/274_Paper.pdf} {A
  universal part-of-speech tagset}.
\newblock In \emph{Proceedings of the Eighth International Conference on
  Language Resources and Evaluation ({LREC}'12)}, pages 2089--2096, Istanbul,
  Turkey. European Language Resources Association (ELRA).

\bibitem[{Platanios et~al.(2019)Platanios, Stretcu, Neubig, Pocz{\'o}s, and
  Mitchell}]{platanios2019competence}
Emmanouil~Antonios Platanios, Otilia Stretcu, Graham Neubig, Barnab{\'a}s
  Pocz{\'o}s, and Tom Mitchell. 2019.
\newblock Competence-based curriculum learning for neural machine translation.
\newblock In \emph{Proceedings of the 2019 Conference of the North American
  Chapter of the Association for Computational Linguistics: Human Language
  Technologies, Volume 1 (Long and Short Papers)}, pages 1162--1172.

\bibitem[{Sachan and Xing(2016)}]{sachan-xing-2016-easy}
Mrinmaya Sachan and Eric Xing. 2016.
\newblock \href {https://doi.org/10.18653/v1/P16-1043} {Easy questions first? a
  case study on curriculum learning for question answering}.
\newblock In \emph{Proceedings of the 54th Annual Meeting of the Association
  for Computational Linguistics (Volume 1: Long Papers)}, pages 453--463,
  Berlin, Germany. Association for Computational Linguistics.

\bibitem[{Samuel et~al.(2023)Samuel, Kutuzov, {\O}vrelid, and
  Velldal}]{samuel-etal-2023-trained}
David Samuel, Andrey Kutuzov, Lilja {\O}vrelid, and Erik Velldal. 2023.
\newblock \href {https://aclanthology.org/2023.findings-eacl.146} {Trained on
  100 million words and still in shape: {BERT} meets {B}ritish {N}ational
  {C}orpus}.
\newblock In \emph{Findings of the Association for Computational Linguistics:
  EACL 2023}, pages 1954--1974, Dubrovnik, Croatia. Association for
  Computational Linguistics.

\bibitem[{Schluter and Varab(2018)}]{schluter-varab-2018-data}
Natalie Schluter and Daniel Varab. 2018.
\newblock \href {https://doi.org/10.18653/v1/D18-1534} {When data permutations
  are pathological: the case of neural natural language inference}.
\newblock In \emph{Proceedings of the 2018 Conference on Empirical Methods in
  Natural Language Processing}, pages 4935--4939, Brussels, Belgium.
  Association for Computational Linguistics.

\bibitem[{Soviany et~al.(2022)Soviany, Ionescu, Rota, and
  Sebe}]{soviany2022curriculum}
Petru Soviany, Radu~Tudor Ionescu, Paolo Rota, and Nicu Sebe. 2022.
\newblock Curriculum learning: A survey.
\newblock \emph{International Journal of Computer Vision}, pages 1--40.

\bibitem[{Stratos et~al.(2016)Stratos, Collins, and
  Hsu}]{stratos-etal-2016-unsupervised}
Karl Stratos, Michael Collins, and Daniel Hsu. 2016.
\newblock \href {https://doi.org/10.1162/tacl_a_00096} {Unsupervised
  part-of-speech tagging with anchor hidden {M}arkov models}.
\newblock \emph{Transactions of the Association for Computational Linguistics},
  4:245--257.

\bibitem[{Wang et~al.(2019)Wang, Pruksachatkun, Nangia, Singh, Michael, Hill,
  Levy, and Bowman}]{wang2019superglue}
Alex Wang, Yada Pruksachatkun, Nikita Nangia, Amanpreet Singh, Julian Michael,
  Felix Hill, Omer Levy, and Samuel Bowman. 2019.
\newblock {SuperGLUE}: A stickier benchmark for general-purpose language
  understanding systems.
\newblock \emph{Advances in Neural Information Processing Systems}, 32.

\bibitem[{Wang et~al.(2023)Wang, Zhang, Li, and Liu}]{wang2022language}
Yile Wang, Yue Zhang, Peng Li, and Yang Liu. 2023.
\newblock \href {https://openreview.net/forum?id=y7CNId2RnV} {Language model
  pre-training with linguistically motivated curriculum learning}.

\bibitem[{Warstadt et~al.(2023)Warstadt, Mueller, Choshen, Wilcox, Zhuang,
  Ciro, Mosquera, Williams, Paranjabe, Linzen, and
  Cotterell}]{warstadt-et-al-2023-babylm}
Alex Warstadt, Aaron Mueller, Leshem Choshen, Ethan~Gotlieb Wilcox, Chengxu
  Zhuang, Juan Ciro, Rafael Mosquera, Adina Williams, Bhargavi Paranjabe, Tal
  Linzen, and Ryan Cotterell. 2023.
\newblock Findings of the 2023 {B}aby{LM} {C}hallenge: {S}ample-efficient
  pretraining on developmentally plausible corpora.
\newblock In \emph{Proceedings of the 2023 {B}aby{LM} {C}hallenge}. Association
  for Computational Linguistics (ACL).

\bibitem[{Warstadt et~al.(2020{\natexlab{a}})Warstadt, Parrish, Liu, Mohananey,
  Peng, Wang, and Bowman}]{warstadt2020blimp}
Alex Warstadt, Alicia Parrish, Haokun Liu, Anhad Mohananey, Wei Peng, Sheng-Fu
  Wang, and Samuel~R Bowman. 2020{\natexlab{a}}.
\newblock {BLiMP}: The benchmark of linguistic minimal pairs for english.
\newblock \emph{Transactions of the Association for Computational Linguistics},
  8:377--392.

\bibitem[{Warstadt et~al.(2020{\natexlab{b}})Warstadt, Zhang, Li, Liu, and
  Bowman}]{warstadt-etal-2020-learning}
Alex Warstadt, Yian Zhang, Xiaocheng Li, Haokun Liu, and Samuel~R. Bowman.
  2020{\natexlab{b}}.
\newblock \href {https://doi.org/10.18653/v1/2020.emnlp-main.16} {Learning
  which features matter: {R}o{BERT}a acquires a preference for linguistic
  generalizations (eventually)}.
\newblock In \emph{Proceedings of the 2020 Conference on Empirical Methods in
  Natural Language Processing (EMNLP)}, pages 217--235, Online. Association for
  Computational Linguistics.

\bibitem[{Wei et~al.(2021)Wei, Bosma, Zhao, Guu, Yu, Lester, Du, Dai, and
  Le}]{wei-et-al-2021}
Jason Wei, Maarten Bosma, Vincent~Y. Zhao, Kelvin Guu, Adams~Wei Yu, Brian
  Lester, Nan Du, Andrew~M. Dai, and Quoc~V. Le. 2021.
\newblock \href {https://arxiv.org/abs/2109.01652} {Finetuned language models
  are zero-shot learners}.
\newblock arXiv:2109.01652.

\bibitem[{Weizman and Snow(2001)}]{weizman2001lexical}
Zehava~Oz Weizman and Catherine~E Snow. 2001.
\newblock Lexical output as related to children's vocabulary acquisition:
  Effects of sophisticated exposure and support for meaning.
\newblock \emph{Developmental psychology}, 37(2):265.

\bibitem[{Wolf et~al.(2020)Wolf, Debut, Sanh, Chaumond, Delangue, Moi, Cistac,
  Rault, Louf, Funtowicz, Davison, Shleifer, von Platen, Ma, Jernite, Plu, Xu,
  Le~Scao, Gugger, Drame, Lhoest, and Rush}]{wolf-etal-2020-transformers}
Thomas Wolf, Lysandre Debut, Victor Sanh, Julien Chaumond, Clement Delangue,
  Anthony Moi, Pierric Cistac, Tim Rault, Remi Louf, Morgan Funtowicz, Joe
  Davison, Sam Shleifer, Patrick von Platen, Clara Ma, Yacine Jernite, Julien
  Plu, Canwen Xu, Teven Le~Scao, Sylvain Gugger, Mariama Drame, Quentin Lhoest,
  and Alexander Rush. 2020.
\newblock \href {https://doi.org/10.18653/v1/2020.emnlp-demos.6} {Transformers:
  State-of-the-art natural language processing}.
\newblock In \emph{Proceedings of the 2020 Conference on Empirical Methods in
  Natural Language Processing: System Demonstrations}, pages 38--45, Online.
  Association for Computational Linguistics.

\bibitem[{Wu et~al.(2021)Wu, Dyer, and Neyshabur}]{wu2021when}
Xiaoxia Wu, Ethan Dyer, and Behnam Neyshabur. 2021.
\newblock \href {https://openreview.net/forum?id=tW4QEInpni} {When do curricula
  work?}
\newblock In \emph{International Conference on Learning Representations}.

\bibitem[{Yadan(2019)}]{hydra}
Omry Yadan. 2019.
\newblock \href {https://github.com/facebookresearch/hydra} {Hydra -- a
  framework for elegantly configuring complex applications}.
\newblock Github.

\bibitem[{Zhang et~al.(2021)Zhang, Warstadt, Li, and
  Bowman}]{zhang-etal-2021-need}
Yian Zhang, Alex Warstadt, Xiaocheng Li, and Samuel~R. Bowman. 2021.
\newblock \href {https://doi.org/10.18653/v1/2021.acl-long.90} {When do you
  need billions of words of pretraining data?}
\newblock In \emph{Proceedings of the 59th Annual Meeting of the Association
  for Computational Linguistics and the 11th International Joint Conference on
  Natural Language Processing (Volume 1: Long Papers)}, pages 1112--1125,
  Online. Association for Computational Linguistics.

\bibitem[{Zhao et~al.(2023)Zhao, Zhou, Li, Tang, Wang, Hou, Min, Zhang, Zhang,
  Dong et~al.}]{zhao2023survey}
Wayne~Xin Zhao, Kun Zhou, Junyi Li, Tianyi Tang, Xiaolei Wang, Yupeng Hou,
  Yingqian Min, Beichen Zhang, Junjie Zhang, Zican Dong, et~al. 2023.
\newblock \href {https://arxiv.org/abs/2303.18223} {A survey of large language
  models}.
\newblock arXiv:2303.18223.

\end{thebibliography}
\bibliographystyle{acl_natbib}

\newpage

\appendix

\section*{Appendix}

\paragraph{Unsupervised POS-tagging.}
The strict-small track we enter does not allow using any external dataset. This restriction disallows usage of any third-party POS taggers, as these tend to be trained with a supervised corpus. To still be able to use POS information we train our own POS tagger using the unsupervised \texttt{anchor-features} part-of-speech algorithm by \citet{stratos-etal-2016-unsupervised}. This algorithm learns a hidden Markov model (HMM) under the assumption that certain tags are associated with words that have no other tags (the anchor words) and uses additional features to improve the estimation process.

We used the default parameters for this algorithm but learn 30 clusters instead of 12. These clusters are lexicalized, labelled only by the anchor word found for each by the algorithm so must be mapped to POS tags for our usage. Unsupervised POS taggers are typically evaluated by mapping each cluster to the most frequently coinciding gold POS tag. However, since this would be taking advantage of supervised data, we instead map each cluster by inspection, choosing the universal part-of-speech tag \citep{petrov-etal-2012-universal} most representative of the anchor word for each cluster. This mapping is many-to-one, with several clusters mapping to the same tag, but no clusters mapped to ADV (adverb) or X (unknown), suggesting that the unsupervised approach failed to coherently group adverbs into a single cluster.

\begin{table}[h]
\begin{tabular}{lrrr}
\toprule
POS Tag & Precision & Recall & F1 \\
\midrule
NOUN & 0.786 & 0.790 & 0.788 \\
DET & 0.820 & 0.772 & 0.795 \\
CONJ & 0.969 & 0.821 & 0.895 \\
NUM & 0.592 & 0.799 & 0.681 \\
PRON & 0.592 & 0.962 & 0.733 \\   
VERB & 0.816 & 0.823 & 0.819 \\
PRT & 0.501 & 0.701 & 0.584 \\
ADJ & 0.673 & 0.554 & 0.608 \\
ADP & 0.842 & 0.888 & 0.864 \\
PUNC & 0.944 & 0.960 & 0.952 \\
\bottomrule
\end{tabular}
\caption{\label{tbl:unsupervised-pos-performance} Accuracy of our unsupervised POS tagger on a per-tag level.}
\end{table}

We also evaluate how well our POS tagger predicts POS tags, compared to the  supervised POS tagging system that is part of the NLTK Python package \cite{bird2009natural}. Table \ref{tbl:unsupervised-pos-performance} summarizes these results. Interestingly, we observe a large difference in our ability to correctly predict different types of POS tokens.

\begin{figure}[h]
\centering
\includegraphics[width=0.5 \textwidth]{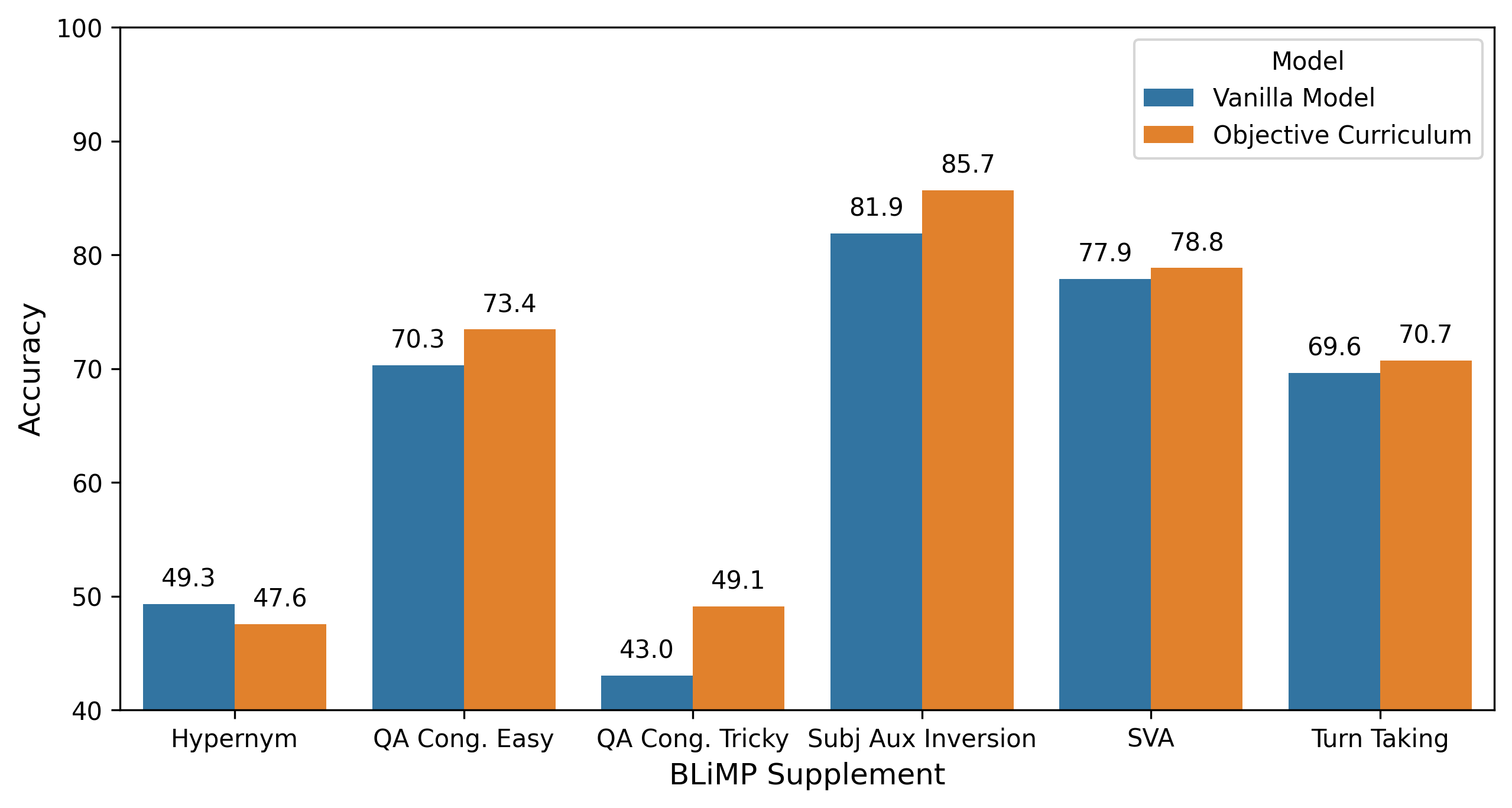}
\caption{\label{fig:baseline_obj_cl_blimp_supp} Comparison between our vanilla model and the best objective curriculum learning setting on the BLiMP supplementary tasks.}
\end{figure}

\begin{figure}[h]
\centering
\includegraphics[width=0.5 \textwidth]{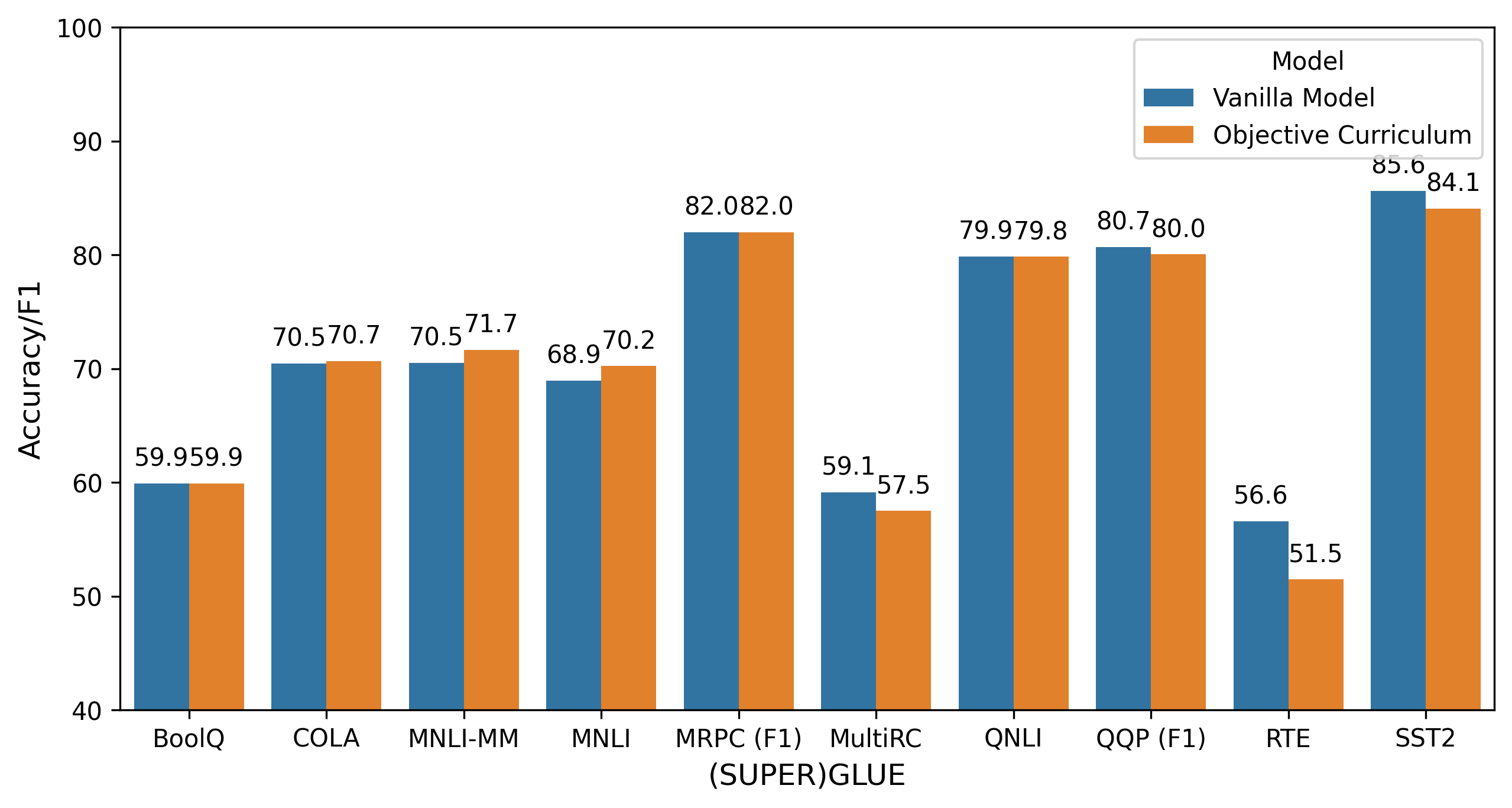}
\caption{\label{fig:baseline_obj_cl_superglue} Comparison between our vanilla model and the best objective curriculum learning setting on the (Super)GLUE tasks.}
\end{figure}

\paragraph{Objective curriculum models on BLiMP Supplement and (Super)GLUE.} Figures \ref{fig:baseline_obj_cl_blimp_supp} and \ref{fig:baseline_obj_cl_superglue} compare our small BabyBERTa-style vanilla  model to our best objective-curriculum model -- a multi-task trained model with sequential POS-tag prediction -- on each task in BLiMP Supplement and (Super)GLUE. We find our curriculum-learning (CL) model outperforms our vanilla model on 5/6 tasks in BLiMP Supplement. While on (Super)GLUE, our CL model outperforms our baseline on 4/10 tasks and obtains comparable performance on another 4/10 tasks. This results illustrate the potential to further explore objective-curricula settings.

\begin{figure*}
\centering
\includegraphics[height=9.5cm]{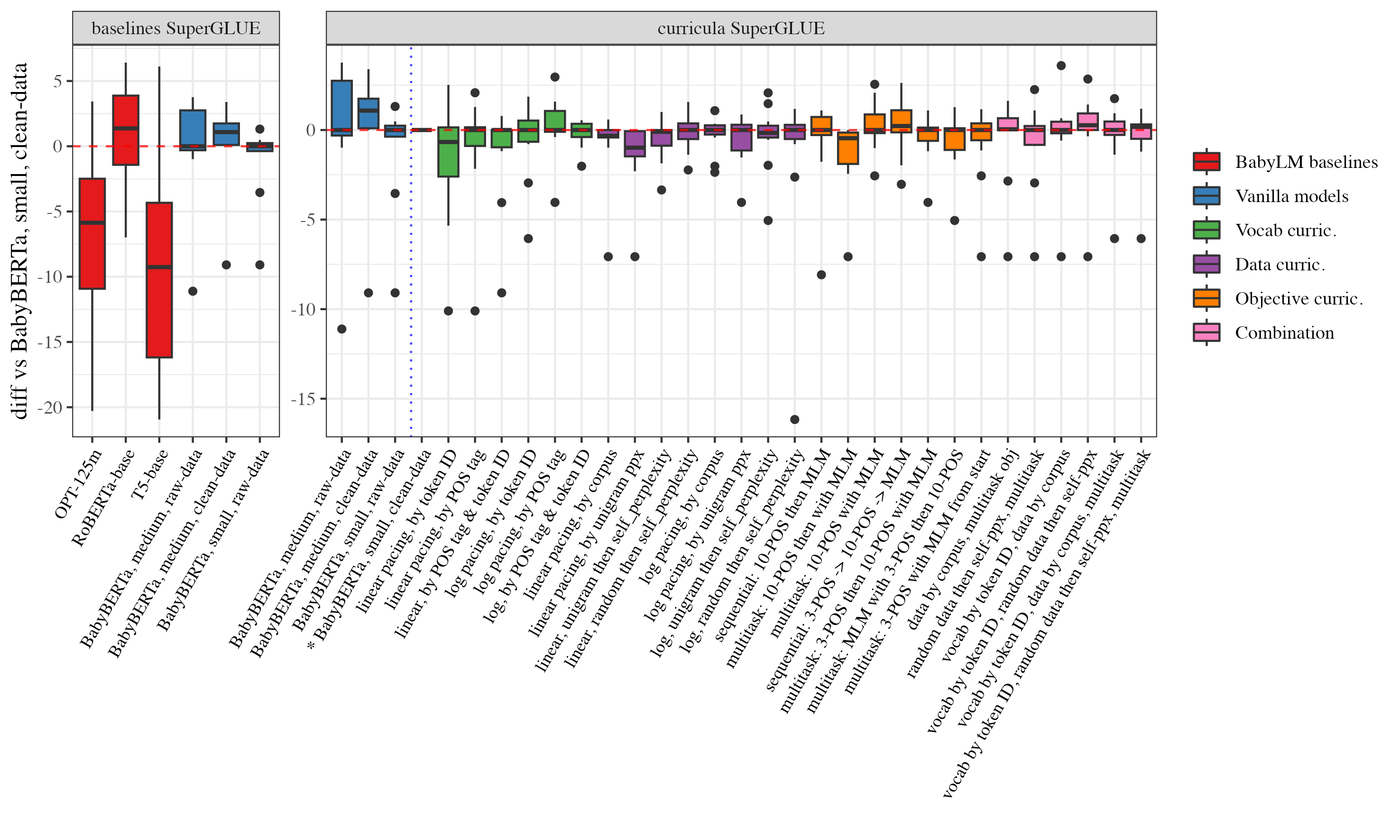}
\caption{\label{fig:glue-boxplots} Comparison of the BabyLM baselines with our BabyBERTa-style vanilla models (left), and our vanilla models against our curriculum learning models (right) -- using BabyBERTa-small trained on clean data as a reference point (asterisked) to show the difference in scores on SuperGLUE tasks. For combination models, all pacing is logarithmic, and `multitask' refers to the 2-task objective curriculum, 10 POS-tags and MLM from the outset. 
}
\end{figure*}

\begin{table*}
\centering
\small
\begin{tabular}{llrrrrr}
\toprule
Pacing & Difficulty         & Perplexity & BLiMP & BLiMP.Supp & (Super)GLUE & MSGS Ambig \\
\midrule
\textsuperscript{\textdagger}Linear & Token ID      &     9.70& 75.09 & \textbf{66.43}      & 68.71            & 68.61    \\
Linear & POS                &   10.17& 72.06 & 63.44      & 69.50            & 66.91    \\
Linear & POS + Token ID     &   10.21& 73.37 & 66.11      & 69.22            & 66.61    \\
Log    & Token ID           &   9.26& 74.97 & 64.63      & 69.94            & 66.82    \\
Log    & POS                &  9.29& 74.12 & 62.06      & \textbf{70.66}            & \textbf{70.52}    \\
Log    & POS + Token ID     &  9.29& 74.74 & 63.62      & 70.29            & 66.42    \\
\midrule
Vanilla Model &  &  \textbf{9.21} &  \textbf{75.48} & 65.34 & 70.47 & 68.30 \\
\bottomrule
\end{tabular}
\caption{\label{tbl:result-vocab-cl} Results for vocabulary curriculum models (Section \ref{subsec:vocab-cl}). All models score above 90 in the MSGS Control tasks. \textsuperscript{\textdagger} indicates the model we submitted to BabyLM, `CLIMB-tokens'. }
\end{table*}

\begin{table*}
\centering
\small
\begin{tabular}{llrrrrr}
\toprule
Pacing & Difficulty         & Perplexity & BLiMP & BLiMP.Supp & (Super)GLUE & MSGS Ambig \\
\midrule
Linear & Source         &    10.41& 73.32 & 61.99      & 69.68      & 66.22    \\
Linear & Unigram ppx     &       12.51& 72.45 & 61.67      & 69.10           & 66.90    \\
Linear & Unigram + model ppx &  11.88& 72.62 & 62.57      & 69.86           & 66.64    \\
Linear & Random + model ppx  &  10.82& 71.88 & 63.10      & 70.37            & 67.48    \\
\textsuperscript{\textdagger}Log    & Source       &      \textbf{9.21}& \textbf{75.87} & 64.29      & 70.20           & \textbf{70.99}    \\
Log    & Unigram ppx      &     9.39& 75.03 & 63.78      & 69.90            & 66.69    \\
Log    & Unigram + model ppx &  9.35& 74.83 & 64.24      & 70.09           & 66.89    \\
Log    & Random + model ppx  &  \textbf{9.21}& 75.81 & 63.03      & 68.93         & 66.64    \\
\midrule
Vanilla Model & &\textbf{9.21}  & 75.48 & \textbf{65.34} & \textbf{70.47} & 68.30 \\
\bottomrule
\end{tabular}
\caption{\label{tbl:result-data-cl} Results for data curriculum models (Section \ref{subsec:data-cl}). All models score above 92 in the MSGS Control tasks. \textsuperscript{\textdagger} indicates the model we submitted to BabyLM, `CLIMB-data-split'. }
\end{table*}

\begin{table*}
\centering
\small
\begin{tabular}{l lll | rrrrr}
\toprule
 & \multicolumn{3}{c}{Task duration (\% of training steps)} & & & & \\
Task Order & 3 POS                                & 10 POS      & MLM        & PPX & BLiMP & BLiMP.Supp & (Super)GLUE & MSGS Ambig \\
\midrule
Sequential       & --       & 0 - 12.5    & 12.5 - 100 & 9.58  & 73.87 & 62.98      & 69.85       & 66.70    \\
Multitask      & --       & 0 - 100     & 12.5 - 100 & 9.78   & 74.60 & 62.17     & 69.12       & 66.64    \\
Multitask      & --       & 0 - 100     & 0 - 100    & 9.30  & \textbf{75.82} & 65.77     & \textbf{70.74}       & 66.58    \\
Sequential       & 0 - 6.25 & 6.25 - 12.5 & 12.5 - 100 & 9.49  & 74.03 & 63.02      & 70.71       & 66.93    \\
Multitask      & 0 - 6.25 & 6.25 - 100  & 12.5 - 100 & 9.72  & 73.68 & 63.89     & 70.07       & 67.00    \\
\textsuperscript{\textdagger}Multitask      & 0 - 6.25 & 6.25 - 100  & 0 - 100    &  9.30 & 74.80 & \textbf{67.55}      & 69.89       & 67.65    \\
Multitask      & 0 - 100  & --          & 0 - 100   & 9.25  & 74.48 & 63.98     & 69.77       & 67.72    \\
\midrule
Vanilla Model & &  & & \textbf{9.21}  & 75.48 & 65.34 & 70.47 & \textbf{68.30} \\
\bottomrule
\end{tabular}
\caption{\label{tbl:result-obj-cl} Results for objective curriculum models (Section \ref{subsec:objective-cl}). All models score above 94 in the MSGS Control tasks. Task duration defines when an objective function was active during training, as a percentage of the total number of training steps. \textsuperscript{\textdagger} indicates the model we submitted to BabyLM, `CLIMB-multitask'. }
\end{table*}

\begin{table*}
\centering
\small
\begin{tabular}{lll|rrrrr}
\toprule
Vocab Curric.\ & Data Curric.\ & Obj. Curric.\ & PPX & BLiMP & BLiMP.Supp & (Super)GLUE & MSGS Ambig  \\
\midrule
-- & Source & Multitask &                           9.29& 74.06 & 64.06 & 70.02 & 66.90 \\
-- & Random + model ppx & Multitask &                9.44& 75.89 & 64.63 & 69.72 & 67.78 \\
Token ID & Source & -- &                     
 9.27& 75.89 & 64.62 & 70.24 & 67.90 \\
Token ID & Random + model ppx & -- &
    9.30& 75.88 & \textbf{65.79} & 70.42 & 66.63 \\
Token ID & Source & Multitask &
     9.22 & 74.86 & 62.82 & 70.09 & 66.68 \\
Token ID & Random + model ppx & Multitask & 9.46& \textbf{75.92} & 63.68 & 69.98 & \textbf{71.30} \\
\midrule
Vanilla Model & & & \textbf{9.21} & 75.48 & 65.34 & \textbf{70.47} & 68.30 \\
\bottomrule
\end{tabular}
\caption{\label{tbl:result-combination-cl} Results for the combination curriculum models. The multitask objective curriculum refers to the 2-task 10-POS and MLM model shown in Table~\ref{tbl:result-obj-cl}. }
\end{table*}

\begin{table*}
\centering
\small
\begin{tabular}{llrrrrr}
\toprule
Type              & Model    & PPX   & BLiMP & BLiMP.Supp & (Super)GLUE & MSGS Ambig \\
\midrule
Official Baseline & OPT-125m         & --    &   63.16 & 55.08 & 63.38 & 69.22 \\
                  & RoBERTa-base      & --  &   
                69.84 & 50.52 & 71.42 & 70.25 \\
                  & T5-base         & --    &   58.27 & 47.55  & 60.93 & 68.55 \\
\midrule
Vanilla Models    &CLIMB-base (medium)   & 9.01   & 75.66 & 66.13 & 70.75 & 67.62 \\
                  & CLIMB-base-small & 9.21  & 75.48 & 65.34 & 70.47 & 68.30 \\
                  & CLIMB-raw (medium)   &  8.47   & 77.97 & 66.16 & 70.63 & 69.44 \\
                  & CLIMB-small-raw  & 8.64  & 76.42 & 64.60 & 69.46 & 70.65 \\
                & \emph{large-100M}      & 4.35      &   81.03 & 75.56 & 72.93 & 74.17 \\
\midrule
Vocab Curriculum          & CLIMB-tokens   &  9.70  & 75.09 & 66.43  & 68.71 & 68.61 \\
Data Curriculum           & CLIMB-data-split & 9.21 & 75.87 & 64.29  & 70.20 & 70.99 \\
Objective Curriculum      & CLIMB-multitask & 9.30 & 74.80 & 67.55  & 69.89 & 67.65 \\
\bottomrule
\end{tabular}
\caption{\label{tbl:submission-comparison} Comparison between the official shared task baselines, our BabyBERTa-style vanilla models, and our submitted curriculum learning models on the main evaluation tasks: BLiMP, (Super)GLUE, and MSGS. Our *small and *medium models are defined in Section \ref{subsec:baseline}. All models are trained on pre-processed data except for those labelled with *-raw, which are trained on mostly unprocessed data (except we join the input sentences). The `large-100M' model was a larger BabyBERTa-style model trained on the 100M BabyLM training set (all others have been trained on the 10M dataset available in the \textsc{strict-small} track). }
\end{table*}

\begin{table*}
    \centering
    \small
    \begin{tabular}{lc}
    \toprule
         Parameter& Value\\
    \midrule
         Layer Norm EPS& 1e-5 \\
         Tie Word Embeddings & False \\
         Learning Rate & 0.001 \\
         Optimizer & AdamW \\
         Scheduler Type & Linear\\
         Max Steps & 400,000 \\
         Warm-up Steps & 100,000\\
         Per Device Batch Size & 32 \\
    \bottomrule
    \end{tabular}
    \caption{Hyperparameter settings which are constant across our vanilla models described in \ref{subsec:baseline}. Table \ref{tbl:baseline-size-comparison} reports variations to the architectures to create the `small', `medium' and `large' versions of the vanilla model. Where values are not reported, they may be assumed to be default values.}
    \label{tbl:baseline_hyperparams}
\end{table*}
\begin{figure*}
\centering
\includegraphics[width=.75\textwidth]{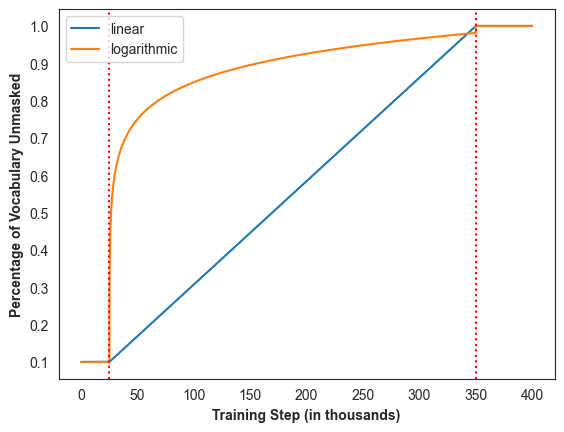}
    \caption{Illustration of the linear and logarithmic pacing functions used in our vocabulary curriculum experiments. The red dotted lines denote the curriculum regime, during which the percentage of unmasked words available to the model grows according to the respective function.}
    \label{fig:pacing_fn}
\end{figure*}
\begin{table*}
    \centering
    \small
    \begin{tabular}{llc}
    \toprule
         Type & Model & Training Time \\
    \midrule
         Vanilla Models & CLIMB-small-raw & 12h \\
         & CLIMB-raw (medium) & 17h40m \\
    \midrule
         Data Curriculum & Log Source & 12h30m \\
         & Log Random + model ppl & 17h10m \\
         Objective Curriculum & Sequential All POS & 11h40m \\
         & Multitask All POS & 15h30m \\
         Vocabulary Curriculum & Linear POS & 11h50m \\
         & Log Token ID & 12h10m \\
    \midrule
        Combination & Log Data Split + Log Token ID & 12h30m \\
        & Log Random + model ppl + Log Token ID & 17h10m \\
    \bottomrule
    \end{tabular}
    \caption{Compute required to train our models. We report the model with the shortest and longest runtime for each experiment type. Each model is trained for 400,000 steps with 4 A100 GPUs.}
    \label{tbl:compute}
\end{table*}
\end{document}